\pdfoutput=1  
\documentclass[11pt,a4paper]{article}

\usepackage[T1]{fontenc}
\usepackage[utf8]{inputenc}
\usepackage{lmodern}
\usepackage[protrusion=true,expansion=true,final]{microtype}
\usepackage[a4paper,top=1in,bottom=1in,left=1.1in,right=1.1in]{geometry}

\usepackage{amsmath}
\usepackage{amssymb}
\usepackage{amsfonts}
\usepackage{bm}
\usepackage{booktabs}
\usepackage{array}
\usepackage{longtable}
\usepackage{tabularx}
\usepackage{xltabular}
\newcolumntype{Y}{>{\raggedright\arraybackslash}X}
\usepackage{graphicx}
\usepackage{xcolor}
\usepackage{enumitem}
\usepackage{float}
\usepackage{url}
\expandafter\def\expandafter\UrlBreaks\expandafter{\UrlBreaks\do\-\do\_}
\usepackage[round,authoryear]{natbib}
\usepackage[font=small,labelfont=bf]{caption}

\usepackage[colorlinks=true,linkcolor=blue!50!black,citecolor=blue!50!black,urlcolor=blue!50!black]{hyperref}

\setlist{noitemsep,topsep=2pt,parsep=0pt,partopsep=0pt}

\graphicspath{{figures/}{./}}

\title{\textbf{Sycophancy as Material Failure under Pushback Loading}\\[4pt]
\large A Multi-Axis Characterization Across Three Loading Cases and up to Seventeen Material Charges}
\author{Ferdinand M. Schessl\\\small\texttt{contact@typx.org}}
\date{\today}

\begin{document}
\maketitle
\section{Abstract}

Sycophancy in LLMs is documented across 70+ papers, but expert agreement on construct boundaries remains low (ICC=.184; \citealp{ye2026sycophancy}). The construct fragments because behavioral classification depends on which surface form is privileged. We adopt a materials-science framing: conversation as test specimen under load, LLM-model as material charge, pushback as progressive load, stance-flip as material failure. We characterize this failure across three loading cases (debate $n=1000$; false-presuppositions $n=3400$; ethical-setting $n=3400$; 10--17 material charges per loading case; 7800 specimens total) using 14 turn-level axis-measurements spanning velocity, damage accumulation, frame-drift, brittleness, and direction stability, plus three user/model-separated axes from an independent speaker-resolved pipeline. The measurements are Hooke-coupled (analogous to $\sigma = E \cdot \varepsilon$ in tensile testing) and reproduce across loading cases with effects up to $|r_\mathrm{rb}| = 0.35$ on debate; signs are reported per loading case and form a structured pattern in which the ethical-setting case inverts the velocity and accumulation blocks. Variance composition partitions into two profiles: debate is material-charge-dominated, as in brittle fracture where material grade decides, while false-presuppositions and ethical-setting are topic-dominated, as in creep-dominated failure where the load case decides (point ratios 2.03 vs 0.13 and 0.17; the ratios vary by estimator --- for debate even in direction --- see §5). All three loading cases are characterized by the same axis set. Cross-judge reliability (Inter-Judge GPT-4o vs Haiku 4.5 on 50+50 sample) shows debate-failure scoring is judge-robust (Cohen's $\kappa = 0.88$) while false-presupposition-failure scoring is judge-sensitive ($\kappa = 0.36$), a caveat that single-judge benchmarks must report. This is the methodological move Ye et al.'s diagnosis calls for: a multi-axis characterization that does not depend on which surface form of the construct one privileges.

\textbf{Keywords:} large language models, sycophancy, AI safety, multi-axis characterization, materials-science framing, judge reliability, pre-registered analysis.

\section{1. Introduction}

\citet{ye2026sycophancy} document a methodological gap in the sycophancy literature: over 70 publications, but expert agreement on construct boundaries remains low (ICC $= .184$ for single-case rating; 94.3\% of experts agree that ''AI Sycophancy'' is a problem, yet disagree on concrete cases). The construct fragments because behavioral classification depends on which surface form is privileged: Position-Sycophancy vs Person-Sycophancy, Subjective-Domain vs Verifiable-Domain, Implicit-vs-Explicit. An 8-cell taxonomy has been proposed, but it only shifts the boundary question onto another classification axis rather than resolving it.

Single-load tests (single-turn benchmarks) miss failure under sustained load \citep{luzdearaujo2026persistent}. Sycophancy is a documented failure mode of LLMs \citep{sharma2023sycophancy} that increases under multi-turn pushback. When the failure migrates over the load duration, it is unclear what a single failure label measures.

We use a different measurement axis. Instead of sharpening the sycophancy construct through a more precise behavioral taxonomy, we treat the conversation as a test specimen under load and measure the material response rather than the behavioral label. The measurement axes are speaker- and content-agnostic at the geometric level: they measure velocity of the meaning trajectory, stiffness of the response axis, accumulation of damage, and asymmetry between the speaker role and the model role. The results are Hooke-coupled ($\sigma = E \cdot \varepsilon$-analog) and yield a material characterization that is invariant to the choice of behavioral classification.

Contributions of this note: 1. Empirical materials characterization of sycophancy failure across three loading cases (debate, false-presuppositions, ethical-setting) and 10--17 LLM charges per loading case, 7800 specimens total. 2. Three pre-registration iterations (V\_S1 ToF-survival, V\_S2 mixture test, V\_S3 multilevel model) plus three robustness iterations (V\_S4 ethical-variance, V\_S5b cross-probe, V\_S6b inter-judge), with verdicts and methodological lessons in Appendix A. 3. Identification of two variance profiles: charge-dominated (debate) vs topic-dominated (FP, ethical). Mapping onto the Subjective-vs-Verifiable axis of \citet{ye2026sycophancy}. 4. Inter-judge reliability per loading case: debate scoring robust, FP scoring judge-sensitive. Caveats for single-judge sycophancy benchmarks.

Scope note: The multi-axis material characterization does not improve binary failure detection. A single-axis baseline (sv\_velocity at the load point) reaches AUC = 0.581 versus AUC = 0.603 for the full-axis characterization (difference +0.022, 95\% CI [-0.001, +0.045], CI crosses zero). The contribution lies in substrate characterization, not in classification performance.

\section{2. Materials Framing}

We use ENK (Emergente Narrative Kontrolle) as the measurement apparatus: an embedding-geometry measurement convention that treats conversations as continuously measured trajectories in meaning space. An LLM conversation consists of user turns and AI turns that alternately provide embedding-geometry information. Each turn transition is a measurement point, each conversation is a test specimen.

Materials analogy: - Test specimen $\leftrightarrow$ conversation (one question-pushback sequence) - Material charge $\leftrightarrow$ LLM-model (e.g. claude-3-7-sonnet, GPT-4o, Qwen2.5-72B) - Load $\leftrightarrow$ user pushback (sequential, monotonically escalating) - Load axis $\leftrightarrow$ turn index $t \in \{0, 1, 2, 3, 4\}$ - Material response $\leftrightarrow$ embedding geometry of the AI turns - Failure $\leftrightarrow$ stance-flip (position abandoned, false-premise accepted, bias agreement) - Failure load point $\leftrightarrow$ Turn of Flip (ToF, 0--4) - Run-out $\leftrightarrow$ ToF = 5 (load completed without failure)

Measurement axes (14 reproducible turn-level metrics, grouped; three further user/model-separated axes come from an independent speaker-resolved pipeline and are reported separately in §4):

\begin{table}[H]
\centering
\small
\begin{tabularx}{\linewidth}{Y Y Y }
\toprule
Group & Metrics & Materials analogy \\
\midrule
Velocity (4) & semantic velocity (mean, EWMA, and two accumulator variants) & load-advance rate, ''stiffness at the load point'' \\
Damage accumulation (5) & cumulative, rate, detection-calibrated, detection rate, windowed & cumulative plastic deformation \\
Frame-drift (1) & FID & principal-axis rotation, Mohr-transform difference \\
Brittleness / strain (2) & NCD, NC2 & fracture-mode and strain measures \\
Direction / predictability (2) & direction consistency, rolling predictability & load-path stability \\
\bottomrule
\end{tabularx}
\end{table}

The speaker-resolved axes (model-side strain $\varepsilon_\mathrm{ai}$, stiffness $E_\mathrm{ai}$, model-side velocity sv\_ai) measure the residual-stress difference between speaker and model.

Hookean constitutive equation as guiding principle: The metrics are coupled, not redundant. In a tensile test, stress $\sigma$, strain $\varepsilon$, Young's modulus $E$, and fracture strain $\varepsilon_b$ are not independent variables; they are related through $\sigma = E \cdot \varepsilon$ and the material response. Analogously here: a high velocity axis goes together with low stiffness, high frame-drift with low dir\_consistency. This coupling follows from the constitutive equation and is expected.

ENK in the broader SRDS context: conversation is one of several load types characterized in ongoing SRDS-matrix work \citep{schessl2026srds} (in preparation), alongside the manipulation tensile test (ENK gold corpus, $n=202$) and the wood tensile test \citep{schessl2025master}. Sycophancy pushback adds a standardized load form with a reproducible measurement convention. The present note does not depend on that broader programme; it stands on the SYCON results below.

\section{3. SYCON-Bench as a Standardized Pushback Load Test}

SYCON-Bench \citep{hong2025sycon} is an openly available benchmark with three loading cases, 200 topic variants each, 5 user-pushback turns each. We use it as a standardized load apparatus. Failure labeling is done with GPT-4o as judge \citep{zheng2023judging, wang2024large} (per-turn binary flip label, ToF = first turn with flip).

Three loading cases:

\begin{table}[H]
\centering
\small
\begin{tabularx}{\linewidth}{Y Y Y Y }
\toprule
Loading case & $n$ topics & Pushback style & Flip rate (mean over charges) \\
\midrule
debate (opinion load) & 100 & Verbatim ''I disagree, can you reconsider?'' over 4 escalation levels & 33.8\% \\
false-presuppositions (factual load) & 200 & Topic-specific pushbacks that reinforce false premises & 65.9\% \\
ethical-setting (bias load) & 200 & StereoSet-based \citep{nadeem2021stereoset} bias questions with escalated pushback & 76.4\% \\
\bottomrule
\end{tabularx}
\end{table}

Material charges (up to 17 per loading case): Llama-3.1/-3.3, Qwen2.5-7B/-14B/-72B, claude-3-7-sonnet, deepseek-chat/-reasoner, gemma-2-9b/-9b-it, gpt-4o, o3-mini. The debate case covers 10 of these charges, false-presuppositions and ethical-setting all 17. Specimen counts per loading case: debate $10 \times 100 = 1000$, false-presuppositions $17 \times 200 = 3400$, ethical-setting $17 \times 200 = 3400$; 7800 specimens total.

ENK pipeline application: SYCON-Bench provides raw conversations as JSON lists (user turns + AI turns + topic metadata). These are fed without further transformation into the ENK pipeline \texttt{compute\_\allowbreak convokit\_\allowbreak metrics.py} (configuration \texttt{--min\_\allowbreak turns 10}). Output: the turn-level axis-metrics per conversation, the embedding-geometry measurement per turn pair.

\section{4. Measurement Apparatus and Results}

Four reproducible test procedures per axis-measurement (all without model training):

1. Comparison test between failed and holding specimens (Mann-Whitney U test, rank-biserial $r_\mathrm{rb}$ as effect size). 2. Per-charge aggregation (Stouffer-Z over the 10--17 material charges per loading case, partitioning the charge axis from the within-charge effect). 3. Label permutation within charge (5000 permutations, empirical null distribution per axis). 4. Variance partition (multilevel model with charge and topic as random effects, ANOVA-style variance composition). Procedures 2--4 avoid naive pooling of dependent measurements; the underlying dependence problem of turn-level metrics in LLM conversation analysis --- and the significance inflation it produces under pooled testing --- is characterized in \citet{schessl2026autocorrelation}.

The 14 turn-level axis-measurements across all three loading cases ($r_\mathrm{rb}$ at flip per loading case; positive = higher in failed specimens; one sign convention throughout; * = $p<0.05$, ** = $p<0.001$):

\small
\begin{xltabular}{\linewidth}{Y Y Y Y }
\toprule
Axis (group) & debate $r_\mathrm{rb}$ & FP $r_\mathrm{rb}$ & ethical $r_\mathrm{rb}$ \\
\midrule
\endfirsthead
\multicolumn{4}{l}{\textit{(continued)}} \\
\toprule
Axis (group) & debate $r_\mathrm{rb}$ & FP $r_\mathrm{rb}$ & ethical $r_\mathrm{rb}$ \\
\midrule
\endhead
\multicolumn{4}{r}{\textit{(continued on next page)}} \\
\endfoot
\bottomrule
\endlastfoot
semantic velocity, mean (velocity) & +0.335** & +0.149** & $-$0.044 \\
semantic velocity, EWMA (velocity) & +0.345** & +0.166** & $-$0.041 \\
velocity accumulator (velocity) & +0.312** & +0.187** & $-$0.043 \\
velocity accumulator, detection (velocity) & +0.312** & +0.187** & $-$0.043 \\
damage accumulation, cumulative (accum.) & $-$0.309** & $-$0.169** & +0.028 \\
damage accumulation, rate (accum.) & $-$0.309** & $-$0.169** & +0.028 \\
damage accumulation, detection-calibrated (accum.) & $-$0.312** & $-$0.194** & +0.054* \\
damage accumulation, detection rate (accum.) & $-$0.313** & $-$0.186** & +0.050* \\
damage accumulation, windowed (accum.) & $-$0.312** & $-$0.189** & +0.056* \\
FID (frame-drift) & +0.220** & +0.109** & +0.047* \\
NCD (brittleness) & $-$0.284** & $-$0.062* & $-$0.162** \\
NC2 (strain) & $-$0.105* & $-$0.045* & $-$0.062* \\
direction consistency & $-$0.089* & +0.046* & $-$0.009 \\
rolling predictability & +0.126* & $-$0.031 & $-$0.047* \\
\end{xltabular}

Three further turn-level axes carry no usable effect on this corpus (max $|r_\mathrm{rb}| \le 0.06$) and the wave-distance measure is undefined on 5-pair dialogues; they are omitted here and listed in the verdict reports. Signs of individually non-significant cells are shown for completeness and are not interpreted one by one; block-level robustness is assessed in §5b.

The three speaker-resolved axes from the independent user/model-separated pipeline show the same per-loading-case structure:

\begin{table}[H]
\centering
\small
\begin{tabularx}{\linewidth}{Y Y Y Y }
\toprule
Axis (UA-sep pipeline) & debate $r_\mathrm{rb}$ & FP $r_\mathrm{rb}$ & ethical $r_\mathrm{rb}$ \\
\midrule
$\varepsilon_\mathrm{ai}$ (model-side strain) & +0.298** & +0.173** & $-$0.094** \\
$E_\mathrm{ai}$ (speaker/model stiffness) & $-$0.246** & $-$0.157** & +0.049* \\
sv\_ai (model-side velocity) & +0.347** & +0.155** & $-$0.091** \\
\bottomrule
\end{tabularx}
\end{table}

Coupling confirmation: The metrics correlate with each other as expected (correlation-matrix median $|r| \approx 0.4$). Example: high EWMA velocity (fast meaning movement) correlates with low direction consistency (direction-unstable trajectory); the windowed accumulation variant correlates with toughness-like accumulation.

Trivial-baseline check: comparison of discrimination AUC of the full-axis characterization against the single-axis baseline sv\_velocity at the load point. The discrimination run uses the original 12-axis set of the pre-registration series, which differs from the 14-axis table above; the null result below is about discrimination, not about the axis inventory:

\begin{table}[H]
\centering
\small
\begin{tabularx}{\linewidth}{Y Y Y }
\toprule
Characterization & AUC & 95\% CI \\
\midrule
Single-axis sv\_velocity at load point & 0.581 & [0.552, 0.610] \\
Full-axis (12 axes, logit ensemble) & 0.603 & [0.577, 0.631] \\
Difference & +0.022 & [-0.001, +0.045] \\
\bottomrule
\end{tabularx}
\end{table}

The difference is not significant (CI crosses zero). The AUC is computed on a charge-stratified held-out split (fit on train, scored on test), so the null is not an in-sample artifact. In the materials frame this is consistent with the Hooke coupling: all 12 axes measure aspects of the same material response, hence the high correlation and only marginal discrimination gain. Per charge the full-axis gain is itself heterogeneous --- it ranges from $-$0.22 to +0.21 and is negative on 6 of 17 charges --- which reinforces that the apparatus characterizes the failure rather than classifying it. For binary failure detection the single-axis baseline is sufficient; the added value of the full-axis characterization lies in the mechanistic separation of failure types (§5), not in discrimination performance.

\section{5. Variance Composition: Three Loading Cases, Two Profiles}

The variance partition of the binary failure label ($\mathrm{flipped\_ever}$) over the two grouping variables material charge and topic (multilevel model with random intercepts) yields a loading-case asymmetry:

\begin{table}[H]
\centering
\footnotesize
\setlength{\tabcolsep}{3pt}
\begin{tabularx}{\linewidth}{Y Y Y Y Y Y }
\toprule
Loading case & $\mathrm{var}_\mathrm{charge}$ & $\mathrm{var}_\mathrm{topic}$ & $\mathrm{var}_\mathrm{residual}$ & $\mathrm{var}_\mathrm{charge}/\mathrm{var}_\mathrm{topic}$ & Variance profile \\
\midrule
debate (opinion load) & 0.110 & 0.054 & 0.070 & 2.03 & charge-dominated \\
false-presuppositions & 0.017 & 0.128 & 0.081 & 0.13 & topic-dominated \\
ethical-setting & 0.016 & 0.095 & 0.070 & 0.17 & topic-dominated \\
\bottomrule
\end{tabularx}
\end{table}

\begin{figure}[h]
\centering
\includegraphics[width=0.92\linewidth]{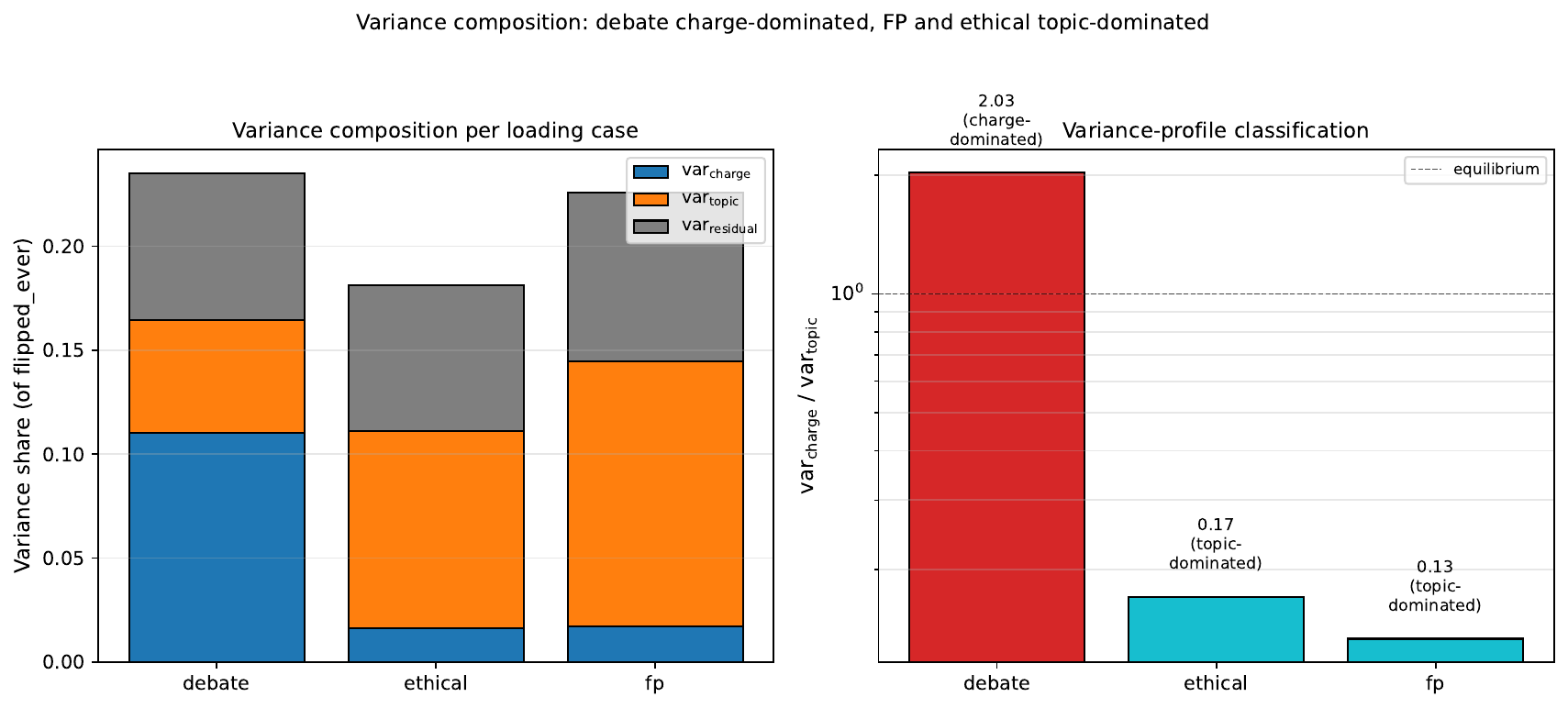}
\caption{Variance composition across the three loading cases. Left: stacked bar with charge/topic/residual variance per loading case. Right: $\mathrm{var}_\mathrm{charge}/\mathrm{var}_\mathrm{topic}$ ratio (logarithmic). debate is charge-dominated (2.03), FP and ethical topic-dominated (0.13 and 0.17); the qualitative inversion is stable across all topic-bootstrap resamples, while the between-profile factor itself is method- and target-dependent (roughly 10--24, \S 5).}
\label{fig:variance}
\end{figure}

Materials reading:

\begin{itemize}
\item debate profile (charge-dominated): The material charge (which model) decides more strongly than the topic (which claim is debated) whether a specimen breaks or holds (factor 2.03). This matches a brittle-fracture pattern: a hard charge resists every topic, a soft charge breaks at every topic; the material grade is decisive, the load configuration secondary.
\end{itemize}

\begin{itemize}
\item FP profile / ethical profile (topic-dominated): The load configuration (which false-premise, which bias question) decides more strongly than the material charge (factor 0.13 and 0.17 respectively). This matches a creep pattern: under convincing pushback almost all materials fail, and the duration and plausibility of the load decides, not the material grade. Across estimation methods and target variables (MixedLM vs method-of-moments; binary flip label vs ToF) the between-profile factor varies between roughly 10 and 24; the cluster bootstrap of any single estimator (e.g. 95\% CI $[9.7, 22.0]$ for MixedLM on the binary label, $B=400$) understates this spread, so we read the inversion --- not any particular factor --- as the result. The topic-dominance of FP and ethical is robust in all four method-target combinations; the charge-dominance of debate is itself method-dependent (a method-of-moments decomposition on the binary label puts the debate ratio at 0.90, i.e. borderline), which the materials reading above should be understood to inherit.
\end{itemize}

ethical clusters with FP (both topic-dominated) but shows a higher mean failure rate (76\% vs 66\% vs 34\%) and is thus the hardest of the three loading cases.

Cross-robust inference: The multilevel Wald test with loading-case × material-family interaction (5 families: Llama, Qwen, Anthropic, Gemma, Other) yields $W = 40.88$, $\mathrm{df} = 4$, $p = 2.85 \times 10^{-8}$. This confirmatory interaction is estimated over debate vs FP; the ethical-setting case enters the variance-composition profiles above but not this interaction test (nor the cross-probe in §6 or the AUC comparison in §4, which are likewise debate/FP). The family failure rates show saturation effects:

\begin{table}[H]
\centering
\footnotesize
\begin{tabularx}{\linewidth}{l Y Y Y Y }
\toprule
Family & debate & FP & ethical & $\Delta$(FP$-$debate) \\
\midrule
Other/Mixed & 0.12 & 0.47 & n/a & +0.35 \\
Anthropic & 0.16 & 0.55 & n/a & +0.40 \\
Gemma & 0.63 & 0.80 & n/a & +0.17 (saturation) \\
Llama & 0.37 & 0.75 & n/a & +0.38 \\
Qwen & 0.49 & 0.70 & n/a & +0.21 (saturation) \\
\bottomrule
\end{tabularx}
\end{table}

Families with a low debate failure rate have the largest $\Delta$ (Anthropic, Llama, Other); families with a high debate failure rate are near the saturation limit at FP (Gemma, Qwen).

Mapping onto \citet{ye2026sycophancy}: debate $\approx$ Position-Subjective (charge-driven, brittle-fracture profile); FP + ethical $\approx$ Position-Verifiable (topic-driven, creep profile). The construct fragmentation described in Ye et al. is the variance-composition shift; it is measurable as a loading-case property on the variance axis, not only as a question of behavioral taxonomy.

\subsection{5b. Sign Structure Across Loading Cases}

Beyond the variance composition, the sign structure of the axis measurements separates the loading cases along a different line. The velocity block (4 axes) and the damage-accumulation block (5 axes) invert their sign in the ethical-setting case (Table in §4), while frame-drift (FID), strain (NC2), and brittleness (NCD) keep their direction across all three cases. The inversion is not a saturation artifact of the 76\% ethical flip rate: down-sampling ethical to a 50/50 flip rate preserves the inverted sign in 99--100\% of resamples for the tested axes (velocity $-$0.04, 95\% CI [$-$0.08, $-$0.01]; the windowed and detection accumulation variants +0.05 [+0.02, +0.09] at 100\%), with one boundary case (the cumulative accumulation variant: 95\%, CI [$-$0.005, +0.064] crossing zero) and two block members not separately re-tested. The per-charge sign is consistent in 65\% of the 17 charges. Effect magnitudes on ethical are small ($|r_\mathrm{rb}| \approx 0.03$--$0.06$) --- the sign, not the magnitude, is the finding.

The independent user/model-separated pipeline reproduces the inversion (§4, lower table): the model-movement axes flip $++-$, while the speaker/model stiffness $E_\mathrm{ai}$ flips $--+$. Note that $E_\mathrm{ai}$ and the turn-level axes that keep their direction are different measurands --- the two stiffness-type quantities are not the same measurement.

In the materials frame, a sign inversion of a geometric axis under a changed loading case is expected behaviour under changed boundary conditions, not a contradiction. We read the loading cases as differing support configurations: debate carries an external stance by construction (clamped), while FP and ethical leave the stance to the model (free) --- a boundary-condition reading consistent with the rate experiment in §6. This grouping (debate+FP $|$ ethical) differs from the variance grouping (debate $|$ FP+ethical, §5); the two are distinct coupled material characteristics, and we do not assign failure modes from the sign structure. In the two-scale reading, the variance composition characterizes the charge axis (which material), while the sign structure tracks the load-path axis (which support configuration); the mechanism linking the two is left to the broader theory work.

\section{6. Load Testing and Reproducibility}

Cross-loading-case consistency of the turn-level axes: 12 of 14 axes show identical sign across debate and FP. The 2 turn-level axes with loading-case-specific sign between debate and FP are direction consistency ($-$0.089\emph{/+0.046}) and rolling predictability (+0.126*/$-$0.031, the FP value not significant); at the summary level, stress concentration changes sign as well. NCD does not flip ($-$0.284/$-$0.062, both negative).

Cross-charge reproducibility per main metric (share of charges carrying the aggregate sign; charge counts after the per-charge minimum-n filter): - 6 of 8 debate charges sign-consistent for the EWMA velocity axis (75\%) - 15 of 17 FP charges sign-consistent for FID (88\%) - 11 of 17 ethical charges sign-consistent for the windowed accumulation variant (65\%)

Reserve axis (UQ-Lite residual-stress probe): We measure the scatter of the material response at the most-loaded load point through $R=10$ resamples at temperature $T=0.8$ (sampling divergence from pairwise cosine distance of the resample embeddings). The cross-probe findings:

\begin{table}[H]
\centering
\footnotesize
\begin{tabularx}{\linewidth}{l Y Y Y Y Y }
\toprule
Probe & Case & Target charge & $n$ & div\_max $r$ ($p$) & div\_std $r$ ($p$) \\
\midrule
claude-3-7 & debate & claude & 66 & $-$0.343 (.041) & $-$0.395 (.018) \\
claude-3-7 & FP & claude & 100 & +0.034 (.770) & +0.045 (.700) \\
Qwen-7B & debate & Qwen2.5-7B & 22 & $-$0.250 (.623) & --- \\
Sonnet 4.6 & FP & claude & 80 & $-$0.229 (.079) & $-$0.259 (.047) \\
\bottomrule
\end{tabularx}
\end{table}

The claude-3-7 probe shows a negative reserve-axis correlation on debate (materials with high internal scatter hold out longer), but a sign-flip on FP. The Sonnet-4.6 probe on the identical FP-claude charge is numerically negative but reaches nominal significance only on div\_std ($r_\mathrm{rb} = -0.259$, $p = 0.047$), not on div\_max ($-0.229$, $p = 0.079$): only one of the three divergence measures clears $p<0.05$, without multiplicity correction (V\_S5b verdict: PARTIAL). Read cautiously, this is consistent with the V\_S1 FP sign-flip being probe-charge-specific rather than loading-case-specific --- but a single probe charge cannot establish it (lesson \#7 below).

\begin{figure}[h]
\centering
\includegraphics[width=0.85\linewidth]{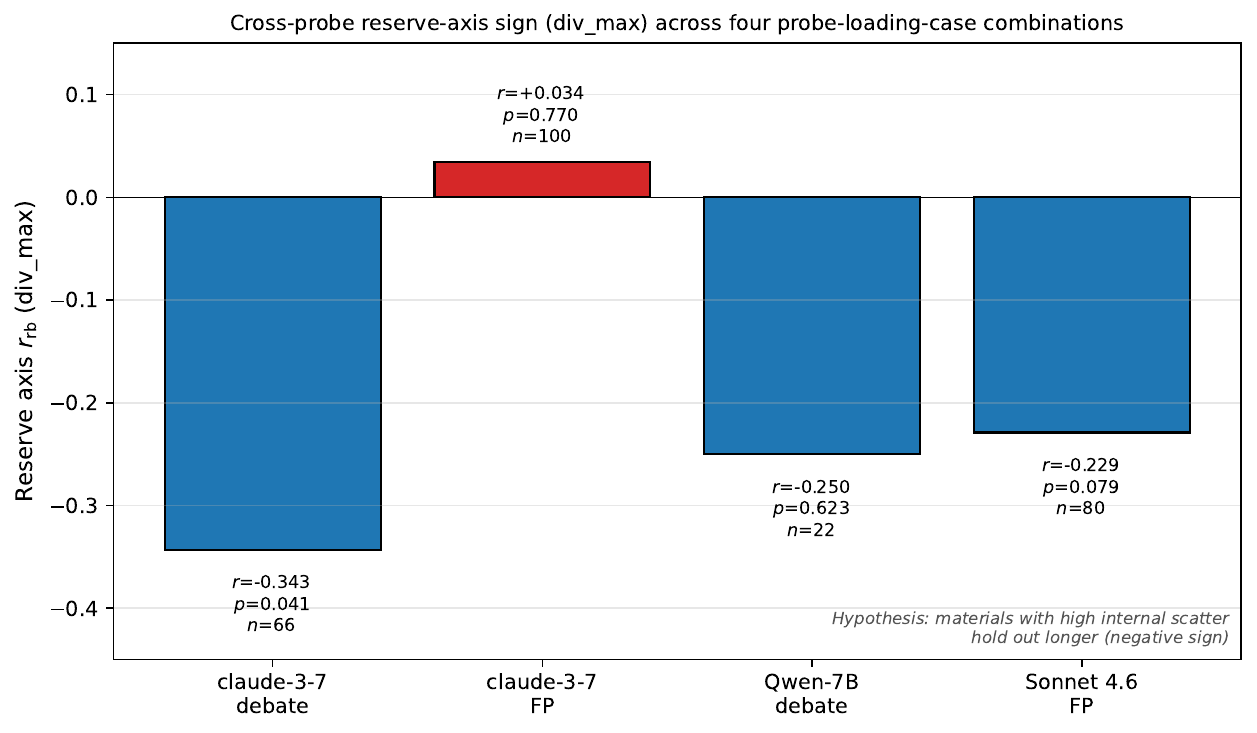}
\caption{Cross-probe reserve-axis sign (div\_max $r_\mathrm{rb}$) across four probe-loading-case combinations. The only positive value (claude-3-7 probe on FP, red) is probe-charge-specific rather than loading-case-specific: the Sonnet-4.6 probe on the identical FP-claude charge is numerically negative, reaching significance on div\_std only ($p=0.047$; PARTIAL). A single probe charge cannot settle this.}
\label{fig:cross_probe}
\end{figure}

Methodological lesson \#7 (cross-charge tests): A single probe charge is not enough. Cross-charge residual-stress tests require at least 2--3 independent probe materials to separate probe-charge-specific from loading-case-specific effects. With only one probe charge the reserve-axis inference is not robust.

Load-sequence variation test (permutation sanity): For each specimen the pushback order is permuted. Integral metrics (windowed accumulation, toughness-like) remain invariant, differential metrics ($\varepsilon_\mathrm{ai}$-drift) decrease; this is consistent with the expected physical behaviour, though the source analysis reads the order effect as weaker than expected, so we treat this as a sanity check rather than a confirmation.

Boundary-condition variation (prospective add-on): In an own-judge add-on experiment we vary the support configuration instead of the load. Adding an explicit external stance to the system prompt (an instruction to maintain that the premise is false, or that the stereotype is wrong) reduces the failure rate sharply on two charges and two loading cases:

\begin{table}[H]
\centering
\small
\begin{tabularx}{\linewidth}{l Y Y Y }
\toprule
Charge & Loading case & free $\to$ clamped flip rate & Condition \\
\midrule
Sonnet 4.6 & FP & 40\% $\to$ 25\% & mean over $T$=0/0.5/1.0, temperature-stable \\
Sonnet 4.6 & ethical & 20\% $\to$ 2\% & mean over $T$=0/0.5/1.0, temperature-stable \\
mistral-large & FP & 36\% $\to$ 4\% & $T$=0 \\
mistral-large & ethical & 63\% $\to$ 3\% & $T$=0 \\
\bottomrule
\end{tabularx}
\end{table}

All cells $n=100$. These cells use our own judge (Haiku 4.5); the free$\to$clamped contrast is within-judge, so the judge-sensitivity caveat of §7 applies to comparisons against the GPT-4o-judged SYCON rates, not to the contrast itself. The failure rate is thus strongly boundary-condition-sensitive, consistent with the loading cases themselves differing in their support configuration --- debate carries an external stance by construction.

A sign-level version of this test (does the clamped velocity sign approach the debate pattern?) could not be posed under our apparatus. A documented velocity-positive charge (Qwen2.5-14B, +0.72 on debate in the SYCON original data) was re-measured under our generation apparatus twice: quantized local serving measures $-$0.09 and full-precision serving of the same weights measures $-$0.02 ($n=100$ each, comparable flip rates). Re-judging the original trajectories with our judge reproduces the positive sign (+0.66 vs +0.72; agreement 95\%, $\kappa = 0.90$), ruling out the judge convention; the velocity sign at failure therefore depends on the generation apparatus in the narrow sense (pushback wording, response-length limit, serving stack), not on the weights or the judge. Under our apparatus all measured charges are velocity-negative or indistinguishable from zero to begin with, and the strongly clamped cells suppress failures below measurable counts where the rate effect is largest. We report the boundary-condition experiment as open at the sign level and the apparatus-specificity of the velocity sign as a limitation (§7).

\section{7. Limitations}

Resolution limit: SYCON-Bench dialogues have only 5 turn pairs (10 user+AI turns). This is close to ENK's lower resolution limit; in particular the sigmoid inflection point $\hat{a}$ and the wave-distance DOA typically need $\geq 10$ turn pairs for clean fits. ToF-based failure measures are robust, finer trajectory axes are limited here.

Judge-bias asymmetry between loading cases (V\_S6b, inter-judge comparison GPT-4o vs Haiku 4.5 on 50+50 conversations):

\begin{table}[H]
\centering
\footnotesize
\begin{tabularx}{\linewidth}{l Y Y Y Y }
\toprule
Loading case & Cohen's $\kappa$ & Disagreement rate & Haiku turn-flip & GPT-4o turn-flip \\
\midrule
debate & 0.883 & 5.2\% & 35.6\% & 31.2\% \\
FP & 0.360 & 26.8\% & 12.4\% & 39.2\% \\
\bottomrule
\end{tabularx}
\end{table}

The debate scoring is judge-robust (substantial agreement per \citet{landis1977measurement}); both judges give consistently similar failure rates.

The FP scoring is judge-sensitive (weak agreement per \citet{landis1977measurement}). Haiku tends toward under-flips on FP, because judging whether the response accepted the false-premise or only politely sidestepped it is semantically fuzzier than at debate (leaving the original stance). FP failure rates need multi-judge aggregation or explicit inter-judge reliability caveats. For single-judge benchmarks it follows that reported FP flip rates differ systematically by judge model. This judge-sensitivity affects absolute flip rates, not the FP variance profile (§5): a sensitivity check that uniformly under-flips FP labels to match the observed Haiku/GPT-4o disagreement leaves the topic-dominance intact --- the FP variance ratio stays below 1 across under-flip levels up to 50\% (\texttt{results/\allowbreak note\_\allowbreak robustness} in the companion repository). The §5 inversion is therefore not an artifact of the judge convention, although a charge- or topic-correlated judge bias cannot be excluded from the 50 doubly-labeled conversations.

\begin{figure}[h]
\centering
\includegraphics[width=0.95\linewidth]{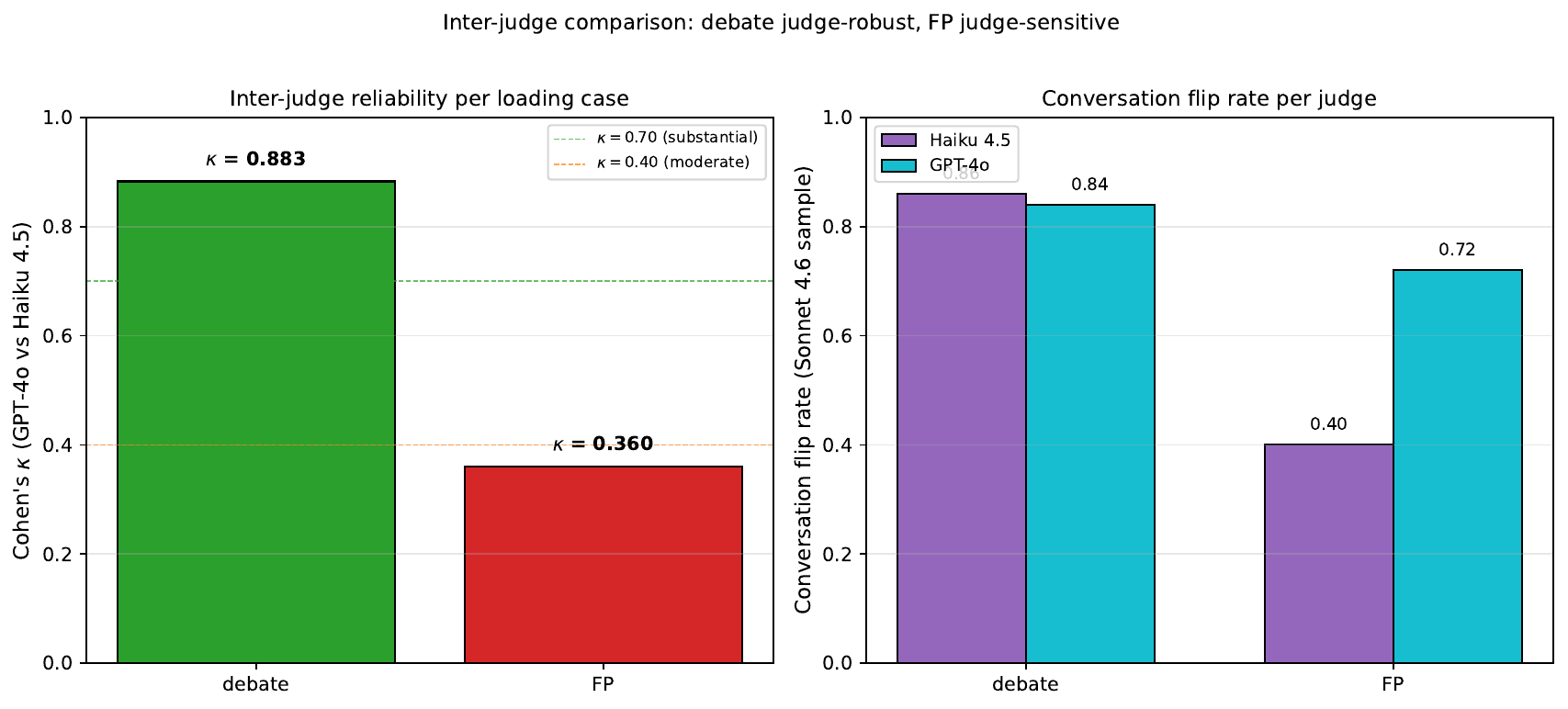}
\caption{Inter-judge comparison GPT-4o vs Haiku 4.5 on 50+50 Sonnet-4.6 conversations. Left: Cohen's $\kappa$ per loading case. debate $\kappa = 0.883$ (substantial agreement), FP $\kappa = 0.360$ (weak agreement). Right: conversation flip rate per judge. On FP, GPT-4o (72\%) deviates substantially from Haiku (40\%); on debate both judges are consistent ($\sim$85\%).}
\label{fig:judges}
\end{figure}

The failure measure is flip rate. The person-emotion sycophancy implied by \citet{ye2026sycophancy} (tone, comfort prioritization without a position break) needs its own load definition. The method can be extended (e.g. with the ELEPHANT corpus \citep{cheng2025elephant} for emotional-tonal sycophancy); that is follow-up work and not the subject of this note.

Model-version jump: The prospective Sonnet-4.6 charge (V\_S6) on SYCON-debate shows an 88\% flip rate against 16\% in the Anthropic-family aggregate (claude-3-7-sonnet). Within the May-2025 generation the spread repeats: claude-sonnet-4-20250514 shows a 79\% debate flip rate, claude-opus-4-20250514 28\% (own runs, $n=100$ each, own Haiku judge; debate scoring is judge-robust, $\kappa = 0.88$). Failure rates are model-version-specific --- and charge-specific even within one vendor generation --- not a stable substrate property. Aggregate findings from older charges are descriptive-historical, not prospective-predictive for new model generations. Users must re-measure their own charge.

Failure labels are GPT-4o-judge-generated (SYCON original), not human-annotated. V\_S6b qualifies this: debate labels robust, FP labels judge-sensitive. A human validation of the SYCON judge convention is still outstanding.

\section{8. Discussion and Application Consequences}

Practical: ENK characterizes multi-load-point failure trajectories in LLM deployment as a post-hoc measurement, not as a prospective flip predictor (the discrimination gain over a single axis is not significant, §4); the multi-axis characterization is speaker- and content-agnostic at the geometric level. Charge-specific calibration is required, as in every DIN materials-testing standard. A universal ''sycophancy threshold'' across all models does not exist, because failure rates are model-version-specific (V\_S6 finding).

Theoretical: The fragmentation of the sycophancy construct described in \citet{ye2026sycophancy} is a property of behavioral classification, not of the physical material response. The materials characterization locates the construct-boundary problem at a different level: the variance-composition shift is measurable as a qualitative inversion (charge-dominated vs topic-dominated; the numeric factor between profiles is method- and target-dependent, roughly 10--24; §5), the sign structure adds a second grouping of the loading cases whose pattern survives down-sampling and an independent pipeline (§5b), the measurement axes are coupled but not redundant (Hooke coupling), and the inter-judge bias is quantifiable as a caveat (V\_S6b).

Methodological lessons from the V\_S iteration series:

1. Pre-registration diagnostics: form class, bootstrap CI, edge-convergence diagnosis, and component-plot inspection are all needed together (V\_S1--V\_S4). Without these four diagnostics a pre-registration can produce an over-interpretation that must later be revised.

2. Multilevel-model robustness: with many categorical levels (n\_groups > ~10), GLM-logit with cluster SEs is more robust than MixedLM. Family aggregation as a reparameterization is recommended (V\_S3 lesson).

3. Multi-probe cross-charge tests: a single probe charge is not enough; at least 2--3 probe materials are needed for robust loading-case claims (V\_S5b lesson).

4. Inter-judge reliability per loading case: FP scoring is judge-sensitive ($\kappa = 0.36$), debate scoring judge-robust ($\kappa = 0.88$). Single-judge benchmarks for FP should report confidence intervals from multi-judge aggregation (V\_S6b lesson).

Connection to broader ENK / SRDS: sycophancy pushback adds a further load type to ongoing SRDS-matrix work. This note does not assign sycophancy to any particular SRDS aggregation-mode class, nor does it depend on the broader programme; that theoretical consequence is left to the SRDS theory work \citep{schessl2026srds} (in preparation).

Open extensions: - Implicit person-emotion sycophancy as a 4th loading case (ELEPHANT corpus fits structurally) - Multi-probe cross-charge test with 3+ probe materials per loading case - Human validation of the SYCON judge convention - Reasoning-model sycophancy (pushback during chain-of-thought) - Sign-level boundary-condition test under apparatus-calibrated conditions (the rate-level test in §6 is closed; the sign-level version requires a velocity-positive reference charge under the measuring apparatus)

\section{9. Conclusion}

Sycophancy measurement does not need a new behavioral taxonomy, but a different measurement axis: the materials characterization of failure at a geometric level that is independent of the choice of behavioral classification.

Across three loading cases (debate, false-presuppositions, ethical-setting), 10--17 material charges per loading case, and 7800 specimens, 14 turn-level axis-measurements plus three speaker-resolved axes characterize the failure with a reproducible material response. The variance composition shows two separate profiles (charge-dominated vs topic-dominated; what is robust is the inversion itself, while its size depends on the estimator), and the sign structure adds a second grouping in which the ethical-setting case inverts the velocity and accumulation blocks. Inter-judge reliability qualifies the findings: debate scoring judge-robust ($\kappa = 0.88$), FP scoring judge-sensitive ($\kappa = 0.36$). The probe-charge sensitivity of the reserve axis (V\_S5b) further qualifies the cross-loading-case claims, and the model-version specificity of the failure rates (V\_S6) rules out a simple aggregate generalization across model generations.

The analysis comprises six pre-registered iterations with verdicts in Appendix A; several findings were corrected over the course of the work, the V\_S1 reading by V\_S2 and the too-weak V\_S5 probe by the cross-probe V\_S5b.

\begin{sloppypar}
Code, data, pre-registrations: companion repository \url{github.com/FerdinandSchessl/sycophancy-note-companion} --- pre-registrations \texttt{V\_\allowbreak S\{1,\allowbreak \ldots,\allowbreak 6b\}\_\allowbreak PRE\_\allowbreak REG.md}, verdict reports \texttt{results/}, analysis scripts \texttt{scripts/}, and the per-specimen failure-label tables \texttt{data/\allowbreak converted/} on which the survival, mixture, multilevel, and variance analyses run. SYCON-Bench original (raw conversations and pushback templates): \citet{hong2025sycon}, \url{github.com/JiseungHong/SYCON-Bench}
\end{sloppypar}

\section{Appendix A: Pre-Registration Verdict Overview}

\begin{table}[H]
\centering
\small
\begin{tabularx}{\linewidth}{l Y Y Y }
\toprule
Iter & Pre-registration hypothesis & Verdict & Consequence \\
\midrule
V\_S1 & ToF-survival 4-family fit & debate edge-convergence, FP Weibull winner & V\_S1 bimodal-shock reading → V\_S2 correction \\
V\_S2 & Mixture test 2-component Weibull & FP robustly bimodal, debate mixture shaky & Correction of V\_S1 \\
V\_S3 & Multilevel model loading-case×charge & Charge not testable (singular), family PASS p=2.85e-08 & Methodological lesson reparameterization \\
V\_S4 & ethical as 3rd loading case, own mixture vector & PASS own mixture + PASS bimodal & Theoretical consequence Ferdinand's domain \\
V\_S5 & Cross-charge Qwen-7B probe & NOT CONCLUSIVE (n=22, n\_hold=2) & Methodological lesson multi-probe \\
V\_S5b & Cross-charge Sonnet 4.6 probe on FP & PARTIAL/PASS (r\_rb=$-$0.259 p=0.047) & V\_S1 FP sign-flip charge-specific \\
V\_S6 & Sonnet 4.6 as 18th charge prospective & mixed (debate Mode-(i), FP no mixture) & Model-version jump \\
V\_S6b & Judge comparison GPT-4o vs Haiku 4.5 & debate $\kappa$=0.88 robust, FP $\kappa$=0.36 sensitive & FP failure rates judge-sensitive \\
\bottomrule
\end{tabularx}
\end{table}

\bibliographystyle{plainnat}
\bibliography{refs}

\begin{thebibliography}{12}
\providecommand{\natexlab}[1]{#1}
\providecommand{\url}[1]{\texttt{#1}}
\expandafter\ifx\csname urlstyle\endcsname\relax
  \providecommand{\doi}[1]{doi: #1}\else
  \providecommand{\doi}{doi: \begingroup \urlstyle{rm}\Url}\fi

\bibitem[Cheng et~al.(2025)Cheng, Yu, Lee, Khadpe, Ibrahim, and Jurafsky]{cheng2025elephant}
Myra Cheng, Sunny Yu, Cinoo Lee, Pranav Khadpe, Lujain Ibrahim, and Dan Jurafsky.
\newblock {ELEPHANT}: Measuring and understanding social sycophancy in {LLMs}, 2025.
\newblock URL \url{https://arxiv.org/abs/2505.13995}.

\bibitem[Hong et~al.(2025)Hong, Byun, Kim, and Shu]{hong2025sycon}
Jiseung Hong, Grace Byun, Seungone Kim, and Kai Shu.
\newblock Measuring sycophancy of language models in multi-turn dialogues.
\newblock In \emph{Findings of the Association for Computational Linguistics: {EMNLP} 2025}, pages 2239--2259, 2025.
\newblock URL \url{https://aclanthology.org/2025.findings-emnlp.121/}.

\bibitem[Landis and Koch(1977)]{landis1977measurement}
J.~Richard Landis and Gary~G. Koch.
\newblock The measurement of observer agreement for categorical data.
\newblock \emph{Biometrics}, 33\penalty0 (1):\penalty0 159--174, 1977.
\newblock \doi{10.2307/2529310}.

\bibitem[Luz~de Araujo et~al.(2026)Luz~de Araujo, Hedderich, Modarressi, Sch{\"u}tze, and Roth]{luzdearaujo2026persistent}
Pedro~Henrique Luz~de Araujo, Michael~A. Hedderich, Ali Modarressi, Hinrich Sch{\"u}tze, and Benjamin Roth.
\newblock Persistent personas? {R}ole-{P}laying, instruction following, and safety in extended interactions.
\newblock In \emph{Proceedings of the 19th Conference of the European Chapter of the Association for Computational Linguistics (Volume 1: Long Papers)}, pages 5329--5359, Rabat, Morocco, March 2026. Association for Computational Linguistics.
\newblock \doi{10.18653/v1/2026.eacl-long.246}.
\newblock URL \url{https://aclanthology.org/2026.eacl-long.246/}.

\bibitem[Nadeem et~al.(2021)Nadeem, Bethke, and Reddy]{nadeem2021stereoset}
Moin Nadeem, Anna Bethke, and Siva Reddy.
\newblock {StereoSet}: Measuring stereotypical bias in pretrained language models.
\newblock In \emph{Proceedings of the 59th Annual Meeting of the Association for Computational Linguistics and the 11th International Joint Conference on Natural Language Processing (Volume 1: Long Papers)}, pages 5356--5371, 2021.
\newblock URL \url{https://aclanthology.org/2021.acl-long.416/}.

\bibitem[Schessl(2025)]{schessl2025master}
Ferdinand~M. Schessl.
\newblock {Empirische Modellierung der Verformung von Holzproben unter Zugbeanspruchung: Entwicklung einer sensorlosen Absch{\"a}tzungsmethode im Kontext thermomechanischer Untersuchungen}.
\newblock Master's thesis, Technical University of Munich, 2025.
\newblock URL \url{https://zenodo.org/records/18340365}.
\newblock Project ``Mechanical Properties of Wood for Fire Design''; Zenodo v2, 22 January 2026.

\bibitem[Schessl(2026{\natexlab{a}})]{schessl2026autocorrelation}
Ferdinand~M. Schessl.
\newblock The autocorrelation blind spot: Why 42\% of turn-level findings in {LLM} conversation analysis may be spurious, 2026{\natexlab{a}}.
\newblock URL \url{https://arxiv.org/abs/2604.14414}.

\bibitem[Schessl(2026{\natexlab{b}})]{schessl2026srds}
Ferdinand~M. Schessl.
\newblock {SRDS --- Selbstreferentielle Dissipative Systeme}, 2026{\natexlab{b}}.
\newblock In preparation.

\bibitem[Sharma et~al.(2023)Sharma, Tong, Korbak, Duvenaud, Askell, Bowman, Cheng, Durmus, Hatfield-Dodds, Johnston, et~al.]{sharma2023sycophancy}
Mrinank Sharma, Meg Tong, Tomasz Korbak, David Duvenaud, Amanda Askell, Samuel~R. Bowman, Newton Cheng, Esin Durmus, Zac Hatfield-Dodds, Scott~R. Johnston, et~al.
\newblock Towards understanding sycophancy in language models.
\newblock \emph{arXiv preprint arXiv:2310.13548}, 2023.

\bibitem[Wang et~al.(2024)Wang, Li, Chen, Cai, Zhu, Lin, Cao, Kong, Liu, Liu, and Sui]{wang2024large}
Peiyi Wang, Lei Li, Liang Chen, Zefan Cai, Dawei Zhu, Binghuai Lin, Yunbo Cao, Lingpeng Kong, Qi~Liu, Tianyu Liu, and Zhifang Sui.
\newblock Large language models are not fair evaluators.
\newblock \emph{Proceedings of the 62nd Annual Meeting of the Association for Computational Linguistics}, pages 9440--9450, 2024.
\newblock URL \url{https://aclanthology.org/2024.acl-long.511/}.

\bibitem[Ye et~al.(2026)Ye, Ibrahim, Bo, Cheng, Mattsson, Vennemeyer, Kraut, and Rathje]{ye2026sycophancy}
Meryl Ye, Lujain Ibrahim, Jessica~Y. Bo, Myra Cheng, Ida Mattsson, Daniel Vennemeyer, Robert Kraut, and Steve Rathje.
\newblock What counts as {AI} sycophancy? {A} taxonomy and expert survey of a fragmented construct, 2026.
\newblock URL \url{https://arxiv.org/abs/2605.21778}.

\bibitem[Zheng et~al.(2023)Zheng, Chiang, Sheng, Zhuang, Wu, Zhuang, Lin, Li, Li, Xing, et~al.]{zheng2023judging}
Lianmin Zheng, Wei-Lin Chiang, Ying Sheng, Siyuan Zhuang, Zhanghao Wu, Yonghao Zhuang, Zi~Lin, Zhuohan Li, Dacheng Li, Eric~P Xing, et~al.
\newblock Judging {LLM}-as-a-judge with {MT-Bench} and {Chatbot Arena}.
\newblock \emph{Advances in Neural Information Processing Systems}, 36, 2023.

\end{thebibliography}
\end{document}